\documentclass[oribibl]{llncs} 
\usepackage[width=180mm,left=18mm,paperwidth=215mm,height=240mm,top=18mm,paperheight=279mm]{geometry}

\usepackage{times}
\usepackage{epsfig}
\usepackage{graphicx}
\usepackage{amsmath}
\usepackage{amsfonts}

\usepackage{float}
\usepackage{color}
\usepackage[justification=centering]{subfig}
\usepackage[ruled,vlined]{algorithm2e}

\begin{document}

\title{Animating Through Warping: an Efficient Method for High-Quality Facial Expression Animation}

\author{Zili Yi\qquad Qiang Tang\qquad Vishnu Sanjay Ramiya Srinivasan \qquad Zhan Xu}
\institute{	Huawei Technologies Canada Co. Ltd.} 

\maketitle

\begin{figure}
\centering
  \includegraphics[width=\textwidth]{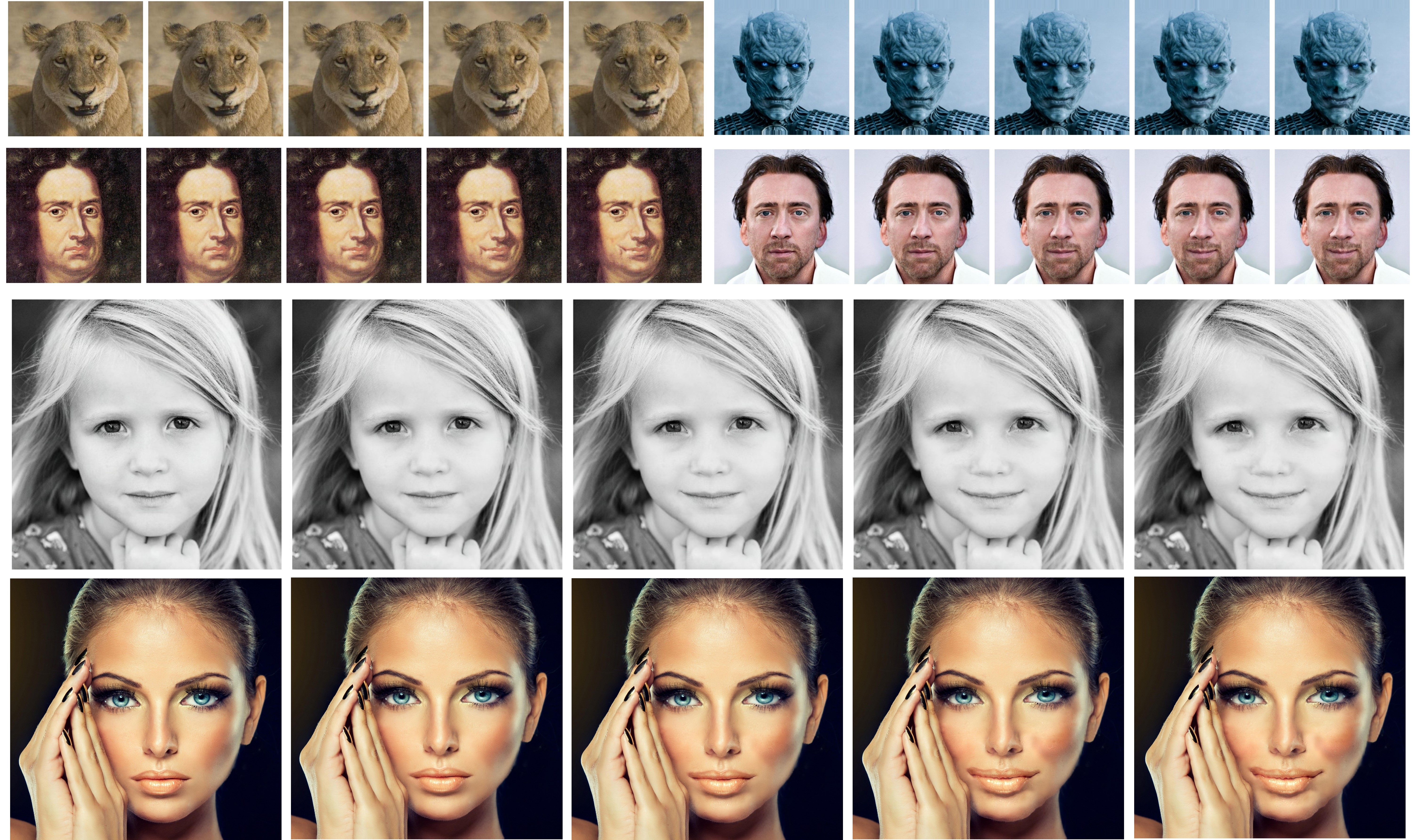}
  \caption{Animation results of sizes ranging from 512 to 4K in the task of making a still face smile. Specifically, images in the first two rows are under 1K, those at the second row are 2K and those in the last row are 4K.}
  \label{fig:teaser}
\end{figure}

\begin{abstract}

Advances in deep neural networks have considerably improved the art of animating a still image without operating in 3D domain. Whereas, prior arts can only animate small images (typically no larger than 512$\times$512) due to memory limitations, difficulty of training and lack of high-resolution (HD) training datasets, which significantly reduce their potential for applications in movie production and interactive systems. Motivated by the idea that HD images can be generated by adding high-frequency residuals to low-resolution results produced by a neural network, we propose a novel framework known as \textbf{A}nimating \textbf{T}hrough \textbf{W}arping (\textbf{ATW}) to enable efficient animation of HD images. 

Specifically, the proposed framework consists of two modules, a novel two-stage neural-network generator and a novel post-processing module known as \textbf{A}nimating \textbf{T}hrough \textbf{W}arping (\textbf{ATW}). It only requires the generator to be trained on small images and can do inference on an image of any size. During inference, an HD input image is decomposed into a low-resolution component(128$\times$128) and its corresponding high-frequency residuals. The generator predicts the low-resolution result as well as the motion field that warps the input face to the desired status (e.g., expressions categories or action units \cite{friesen1978facial}). Finally, the \textbf{ResWarp} module warps the residuals based on the motion field and adding the warped residuals to generates the final HD results from the naively up-sampled low-resolution results. Experiments show the effectiveness and efficiency of our method in generating high-resolution animations. Our proposed framework successfully animates a 4K facial image, which has never been achieved by prior neural models. In addition, our method generally guarantee the temporal coherency of the generated animations. Source codes will be made publicly available.

\end{abstract}

%


\section{Introduction}

Animating a still image of an object or a scene in a controllable, efficient manner enables many interesting applications in the field of image editing/enhancement, movie post-production and human-computer interaction. One popular direction in this domain requires parametric 3D face modeling and then perform animation in 3D domain, e.g., ~\cite{Thies2016}. Approaches in this stream often require post-processing using computer graphics techniques, expensive equipment and significant amounts of labors to produce realistic results. In order to drive down the cost and time required to produce high quality animations, researchers are looking into fully-automatic 2D-based methods using machine learning techniques. Recent years have seen considerable progresses of 2D-based methods in animating faces/heads~\cite{wang2019few,gabbay2019style,songsri2019face,shen2019facial,pumarola2018ganimation,nirkin2019fsgan,sanchez2018triple,wang2019u,wiles2018x2face,hassan2019ganemotion}, human objects~\cite{wang2019few,villegas2017learning,balakrishnan2018synthesizing,zhao2019video,lin2019creating}, cartoon characters\cite{mckenzie2019techniques,sin20192d}, other objects~\cite{siarohin2019animating} and scenes~\cite{wang2019few,vondrick2016generating,endo2019animating,nam2019end,roache2019adding}.

Among these methods, some targeted on animating a still image in a video prediction manner~\cite{vondrick2016generating,villegas2017learning,zhao2019video}, and thus do not need a driving sequence to direct the animating. Some other methods use either exemplars or interpretable attributes to drive the animating. Depending on the application scenario, various forms of driving signals are used, including but not limited to key points~\cite{songsri2019face,gu2019flnet,nirkin2019fsgan,sanchez2018triple,tripathy2020icface,wiles2018x2face,balakrishnan2018synthesizing,geng2018warp}, label maps \cite{wang2019few,zhang2019one}, facial action units\cite{pumarola2018ganimation,tripathy2020icface,hassan2019ganemotion}, skeletons~\cite{wang2019few}, edge maps \cite{wang2019few}, expression categories~\cite{wang2019u,tripathy2020icface}, RGB images~\cite{gabbay2019style,wiles2018x2face}, motion hints~\cite{sin20192d,endo2019animating,siarohin2019animating,siarohin2019first} or speech/audio~\cite{vougioukas2019realistic,wiles2018x2face,hassan2019ganemotion,mckenzie2019techniques}. 

Majority of methods employ an end-to-end neural network to generate one frame at a time and merging a sequence of continuous frames forms an animation. One of the earliest methods is proposed by Shi et al.~\cite{xingjian2015convolutional}, in which they use a ConvLSTM generator to forecast the future rainfall intensity in a local region based on the current rainfall intensity. The generator is trained in a supervised manner with real video sequences.~\cite{villegas2017learning,vondrick2016generating} inherit the idea of training a generator in a supervised manner with real video datasets, and innovatively introduce the adversarial losses \cite{goodfellow2016nips} to improve image quality. In 2018, a novel architecture known as Ganimation~\cite{pumarola2018ganimation} follows an image-to-image translation paradigm, instead of using a sequential model. In particular, they condition the generation of an image on action units (AUs) \cite{friesen1978facial} (a descriptor of facial actions), and produce temporally-coherent animation sequences through continuously altering the AU codes. Different from prior models that are trained with real video datasets, Ganimation is trained with a set of discrete images and weakly-labeled attributes (action units) of those images.~\cite{wiles2018x2face,balakrishnan2018synthesizing,vougioukas2019realistic,gu2019flnet,wang2019few,tripathy2020icface} continue to use the image-to-image translation paradigm, while they employ different forms of driving signals (e.g., key points, edge maps, audio).

Since existing methods employ convolutional layers directly on the raw input, they cannot efficiently handle large images as the memory usage during training and inference could become extremely high when the input size increases. Even assuming the training is feasible, access to large amounts of high resolution training data would be tedious and expensive. Motivated by the idea that HD images can be generated by adding high-frequency residuals to low-resolution results produced by a neural network, we propose a novel \textbf{A}nimating-\textbf{T}hrough-\textbf{W}arping (\textbf{ATW}) framework that enables animation of an HD image with limited resources. The proposed framework inherits the image-to-image translation paradigm and is trained with a set of discrete images with labeled attributes. 

Our proposed  \textbf{ATW} framework consists of two modules, a neural-network-based generator and a \textbf{Res}idual \textbf{Warp}ing (\textbf{ResWarp}) module. Firstly, the raw input face is decomposed into a low-resolution component and high-frequency residuals. The generator processes the low-resolution component and predicts low-resolution animation results as well as a motion field that could warp the input face to the desired status. The \textbf{ResWarp} module then warps residuals based on the up-sampled motion field and add the warped residuals to the up-sampled low-resolution results to generate large sharp results. Since the generator only operates on low-resolution images, the cost of memory and computing time is significantly reduced. Moreover, as the generator can be trained with low-resolution images, the need for HD training datasets is alleviated. In theory, our method can animate a still image of any size. Exemplar results are shown in Figure \ref{fig:teaser}. Contributions of the paper are summarized as bellow:

\begin{itemize}
\item{We propose a novel \textbf{A}nimating-\textbf{T}hrough-\textbf{W}arping (\textbf{ATW}) framework that enables efficient animation of an HD image with satisfying quality. The framework has been successfully verified in terms of the application of one-shot facial expression animation. It successfully animates a 4K face, which is intractable for prior neural models.}

\item{As part of the \textbf{ATW} framework, we propose a novel two-stage generator that predicts a coarse result by warping and then a refined result. In addition, a novel training loss known as refine loss is proposed to force minimum change between the coarse and the refined results, which proved to improve the temporal coherency and image quality of generated image sequences.}

\item{We propose a novel multiscale \textbf{ResWarp} variation that warps image residuals at multiple scales, which brings noticeable improvements over vanilla \textbf{ResWarp} as demonstrated in experiments.}

\end{itemize}

\section{Related work}
\subsection{GANs}
The generator we used in method is a conditional GAN \cite{goodfellow2016nips,mirza2014conditional}, in which the resulting image is conditioned on the input image to be animated and driving signals that define the desired status. The driving signals can have various forms including but not limited to expression categories, action units, key points or edge maps.
\subsection{Predicting motion field for image editing}

In the domain of facial image editing, methods such as \cite{athar2019self} and \cite{wu2019attribute} propose to predict a motion field to warp a face to the target expression. Although our method also predicts a motion field, we targeted on animating rather than image editing, which requires handling of temporal coherency. In addition, in our proposed framework the motion field is used to warp both low-resolution image and high-frequency residual map.

\subsection{Animating still images through warping}

Prior works such as \cite{siarohin2019animating,endo2019animating,zakharov2019few,siarohin2019first} explicitly predict optic flows to facilitate animating of still images. However, there are some aspects that our method differs from these methods. For instance, in \cite{siarohin2019animating,siarohin2019first} motion fields are refined from a sparse motion representation of key points and motion information extracted from the driving video, and in \cite{zakharov2019few} motion fields are also estimated from existing image pairs, whereas we predict motion fields from scratch. \cite{endo2019animating} trains their model with ground-truth optic flows estimated from real videos. However, our model is trained in self-supervised manner with a set of discrete images. Lastly, neither of the three methods explicitly separate the warping of low- and high-frequency components of the target image, thus unable to reenact HD images. The maximum image size experimented are 64$\times$64, 256$\times$256, 640$\times$360 and 512$\times$512 respectively in \cite{siarohin2019animating}, \cite{siarohin2019first}, \cite{endo2019animating} and \cite{zakharov2019few}.

\subsection{Face reenactment/retargeting}

Face reenactment/retargeting aims at generating movements of a target face based on a driving sequence. Typically the target face is provided through a video or a handful of photographs\cite{zhu2017unpaired,yi2017dualgan,wu2018reenactgan,zakharov2019few,korshunova2017fast,wiles2018x2face}. When only one image is provided to define the target face, it is the same as our task. In the experimental section, we will compare our method with face reenactment method such as x2face \cite{wiles2018x2face} and few-shot vid2vid \cite{zakharov2019few}.

\subsection{Video prediction}
Predicting a video from previous frames is an important but challenging task~\cite{finn2016unsupervised,babaeizadeh2017stochastic,denton2018stochastic,denton2018stochastic}. When video prediction is conditioned on the first frame only (e.g., \cite{vondrick2016generating,villegas2017learning,zhao2019video}), it is exactly the same task as animating a still image. However, the task of video prediction does not require a driving video or driving sequence, thus not allowing flexible controllability over generated videos, which is different from our task.

\subsection{Image residuals and Laplacian Pyramid}

The difference between an image and the blurred version of itself represents the high-frequency component of an image. We follow this concept to decompose an image into low- and high-frequency components~\cite{winnemoller2012xdog}. The low-frequency component can be obtained by blurring and down-sampling, while the high-frequency component (i.e. image residuals) can be acquired by subtracting the original image with its blurred version. A more sophisticated way to decompose an image is the Laplacian Pyramid~\cite{burt1983laplacian}, in which an image is down-sampled and decomposed at multiple levels and thus forms a pyramid of scale-disentangled representations. We exploit these techniques in the proposed \textbf{Residual Warping} module.


\begin{figure*}
	\centering
	\includegraphics[width=.9\linewidth]{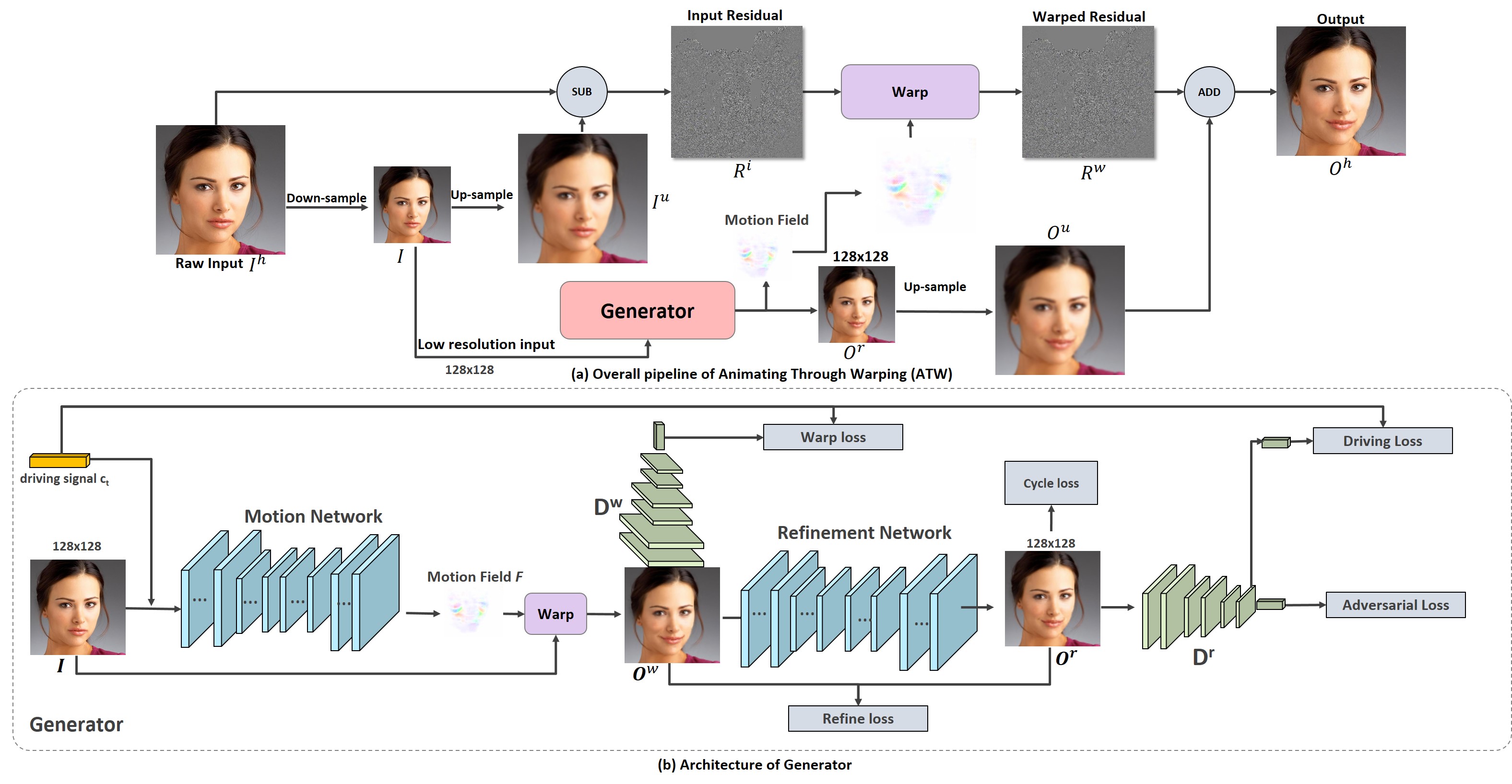}
	\caption{Overall Pipeline of the Proposed Method.\label{fig:method}}
\end{figure*}

\section{Method}

\subsection{Overall pipeline}

Figure~\ref{fig:method} (a) illustrates the overall pipeline of the proposed  \textbf{ATW} framework where the generator is the only trainable module in the framework. Given an HD input image $\textbf{I}^h$, we first down-sample the image to 128$\times$128 and then up-sample it to its original size to obtain a blurry large image $\textbf{I}^u$. The residual image $\textbf{R}^i$ is computed by subtracting $\textbf{I}^u$ from  $\textbf{I}^h$. Note that the height and width of the input image are not necessarily equal but must be multiples of 128, as the input image is evenly divided into 128$\times$128 grids and pixel values in each grid are averaged when down-sampling. The generator takes the low-resolution image $\textbf{I}$ (i.e., 128$\times$128) and generates a manipulated image $\textbf{O}^r$ based on the driving signal (e.g., expression categories, action units). Meanwhile, a motion field is predicted by Motion Network of the generator. The predicted motion field is up-sampled to the same size as the input and then used to warp the residual image $\textbf{R}^i$ to obtain $\textbf{R}^w$. Finally, adding $\textbf{R}^w$ to $\textbf{O}^u$ which is up-sampled from $\textbf{O}^r$ yields a high resolution and sharp output $\textbf{O}^h$. 

Figure~\ref{fig:method} presents the process of generating a single frame. We can generate consecutive frames by either continuously changing the value of driving signals or linearly interpolating the motion field~\cite{pumarola2018ganimation}.

\subsection{Network architecture}

The network architecture of the generator $G$ is shown in Figure~\ref{fig:method} (b). The generator is made up of two sub-networks where the Motion Network predicts a motion field that could warp the input image to the desired status defined by the driving signal, and the Refinement Network refines the warping result and makes it visually more realistic. Generator $G$ takes Image $\textbf{I}$ and driving signal $\textbf{c}_t$ as inputs and predicts a desired image $\textbf{O}^r=G(\textbf{I},\textbf{c}_t)$. The size of inputs and outputs of the generator are expected to be $128\times128$. The driving signal $\textbf{c}_t$ is a vector that defines what is desired in the output, which might have various forms including expression categories and action units. $\textbf{c}_t$ is repeated and tiled to 128$\times$128$\times$d (d is the number of dimensions of the vector) before concatenated to $\textbf{I}$. The predicted motion field indicates the horizontal and vertical displacement of each pixel. Then a coarse result $\textbf{O}^w$ is obtained by warping the input image $\textbf{I}$ with the motion field. The Refinement Network then processes the coarse result $\textbf{O}^w$ and predict a finer one $\textbf{O}^r$. Both the Motion Network and the Refinement Network are designed as an encoder-decoder architecture, in which the input is convolved and down-sampled twice, then processed by bottleneck layers and finally deconvolved and up-sampled twice to the original size. The bottleneck layers are made up of several residual blocks (6 for the Motion Network, 4 for the Refinement Network). 

Instead of leaving the last layer of the Motion Network unbounded, we utilize $\mathbf{tanh}$ as the activation function to limit output bounded within $[-1, 1]$. A factor is then multiplied to estimated motion values to adapt to specific image sizes. We use instance normalization and RELU for each layer in the Motion Network, while use ELU and no normalization layer in the Refinement Network to preserve color consistency between $\textbf{O}^w$ and $\textbf{O}^r$, as RELU and normalization layers are found to deteriorate color consistency~\cite{iizuka2017globally}. Pixel values of all images processed by the network are within range of $[-1, 1]$.
\subsection{Training methodology}
\subsubsection{Training losses}

The training objectives we define consists of five terms: the adversarial loss, the driving loss, the warp loss, the refine loss and the cycle loss: see Figure \ref{fig:method} (b). The adversarial loss forces the distribution of generated images $\textbf{O}^r$ to comply with that of real images; the driving loss is to drive the refined result $\textbf{O}^r$ to satisfy what is desired by the driving signal $\textbf{c}_t$; the warp loss is similar to the driving loss, which pushes the warping result $\textbf{O}^w$ to fit $\textbf{c}_t$; the refine loss minimizes the change of the refined result $\textbf{O}^r$ from the coarse result $\textbf{O}^w$; the cycle loss ensures that the input image $\textbf{I}$ can be recovered from $\textbf{O}^r$ by using its own feature $\textbf{c}_s$ as driving signal.

To facilitate the training of Generator, we used two discriminators, i.e., $\textbf{D}^w$ acting on the warping result $\textbf{O}^w$ and $\textbf{D}^r$ for the refined result $\textbf{O}^r$. Discriminator $\textbf{D}^r$ takes an image as input and predicts two logits: the adversarial logits $\textbf{D}^r_{adv}(.)$ that rates the realism of an image, and the driving logits $\textbf{D}^r_{drv}(.)$ that maps the input image to the corresponding driving signal. As we use the WGAN-GP~\cite{gulrajani2017improved} loss as our adversarial loss, the adversarial loss for Discriminator  $\textbf{D}^r$ is written as:

\begin{equation}
l^{D^r}_{adv} = \mathbb{E}_{\mathbf{I} \sim \mathbb{P}_I, \mathbf{c}_t \sim \mathbb{P}_c} [\mathbf{D}^r_{adv}(\mathbf{I}) -  \mathbf{D}^r_{adv}(G(\mathbf{I}, \mathbf{c}_t)] + \\
\sigma [\| \bigtriangledown_{\hat{\mathbf{I}}} \mathbf{D}^r_{adv}(\hat{\mathbf{I}})\|_2 - 1]^2
\label{eq:dlossr_gan}
\end{equation}
\noindent where $\textbf{D}^r_{adv}(.)$ is the adversarial logits of $\textbf{D}^r$. $\mathbf{I}$, $G(\mathbf{I}, \mathbf{c}_t)$, $\mathbf{\hat{I}}$, are real images, generated images, and the interpolations between them, respectively. $\mathbb{P}_I$, $\mathbb{P}_c$, are distributions of real images $\textbf{I}$ and driving signals $\mathbf{c}_t$ separately. $\sigma$ is the coefficient for the gradient penalty term, typically set to 10.

The driving loss for the discriminator $\textbf{D}^r$ is written as:
\begin{equation}
l^{\mathbf{D}_r}_{drv} = \mathbb{D}(\mathbf{c}_s, \mathbf{D}^r_{drv}(\mathbf{I}))
\end{equation}
where $\mathbb{D}(.)$ measures the disparity between the ground-truth annotation $\mathbf{c}_s$ and the predicted signal vector $\mathbf{D}^r_{drv}(\mathbf{I})$. Depending on the form of driving signals, $\mathbb{D}(.)$ could vary. In our experimental setting, we use Cross Entropy to indicate disparities of expression categories and Mean Squared Error (MSE) to measure distances of action units. 

The training loss for Discriminator $\textbf{D}^r$ is written as:

\begin{equation}
l^{D^r} = \lambda_{drv} l^{D^r}_{drv} + \lambda_{adv} l^{D^r}_{adv}
\label{eq:dlossr}
\end{equation}
where $\lambda_{drv}$ and $\lambda_{adv}$ are coefficients for the driving loss term and adversarial loss term respectively.

Discriminator $\textbf{D}^w$ has only one head and it maps an image to the corresponding driving signal. Therefore, it is only trained with driving loss:
\begin{equation}
l^{\mathbf{D}_w} = \mathbb{F}(\mathbf{c}_s, \mathbf{D}^w(\mathbf{I}))
\label{eq:dlossw}
\end{equation}

The adversarial loss for Generator $G$ is defined as below.
\begin{equation}
l_{adv} = - \mathbb{E}_{\mathbf{I} \sim \mathbb{P}_I, \mathbf{c}_t \sim \mathbb{P}_c} \mathbf{D}^r_{adv}(G(\mathbf{I}, \mathbf{c}_t))
\label{eq:loss_adv}
\end{equation}

We also employ the reconstruction loss (also known as cycle loss\cite{zhu2017unpaired,yi2017dualgan}) to force consistency of the prediction with the original image. We use L1 to compute the loss as it is proved to encourage sharper result. The cycle loss is written as:

\begin{equation}
l_{cyc} = | G(G(\mathbf{I}, \mathbf{c}_t), \mathbf{c}_s) - \mathbf{I} |  \\
\label{eq:loss_rec}
\end{equation}
where $|.|$ is the L1 operation.

The driving loss and warp loss force consistency of outputs with the target driving signal $\textbf{c}_t$. They are written as:

\begin{equation}
l_{drv} = \mathbb{D}(\mathbf{c}_t, \mathbf{D}^r(\mathbf{O}^r))
\label{eq:loss_ctrl}
\end{equation}

\begin{equation}
l_{warp} = \mathbb{D}(\mathbf{c}_t,\mathbf{D}^w(\mathbf{O}^w))
\label{eq:loss_warp}
\end{equation}

The refine loss forces minimum modification of the refined result $\textbf{O}^r$ upon the coarse warping result $\textbf{O}^w$, thus written as:

\begin{equation}
l_{ref} =  | \mathbf{O}^w - \mathbf{O}^r |
\label{eq:loss_refine}
\end{equation}

The proposed refine loss has two benefits: 1, maximize the change brought by warping, so that the motion and appearance are fully decoupled, and thus the motion field is well aligned with the low- and high-frequency components; 2, minimize the color variance caused by refinement to avoid color flicking, which potentially helps improve the temporal coherency of generated image sequences.

The Motion Network and the Refinement Network are trained jointly with a weighted sum of the five losses, as shown in Equation~\ref{eq:gloss}.

\begin{equation}
l_g = \lambda_{cyc} l_{cyc} + \lambda_{adv} l_{adv}  + \lambda_{drv} l_{drv} + \lambda_{warp} l_{warp} + \lambda_{ref} l_{ref}  
\label{eq:gloss}
\end{equation}
\noindent where $\lambda_{rec}$, $\lambda_{drv}$, $\lambda_{warp}$, $\lambda_{adv}$ and $\lambda_{ref}$ are coefficients for each loss terms. In our experiments, we set  $\lambda_{rec}=10.$, $\lambda_{drv}=1.$, $\lambda_{warp}=1.$, $\lambda_{adv}=1.$ and $\lambda_{ref}=10.$ in our final implementation. However, we suggest further optimization of configurations based on specific requirements.

\begin{figure*}
	\centering
	\includegraphics[width=.9\linewidth]{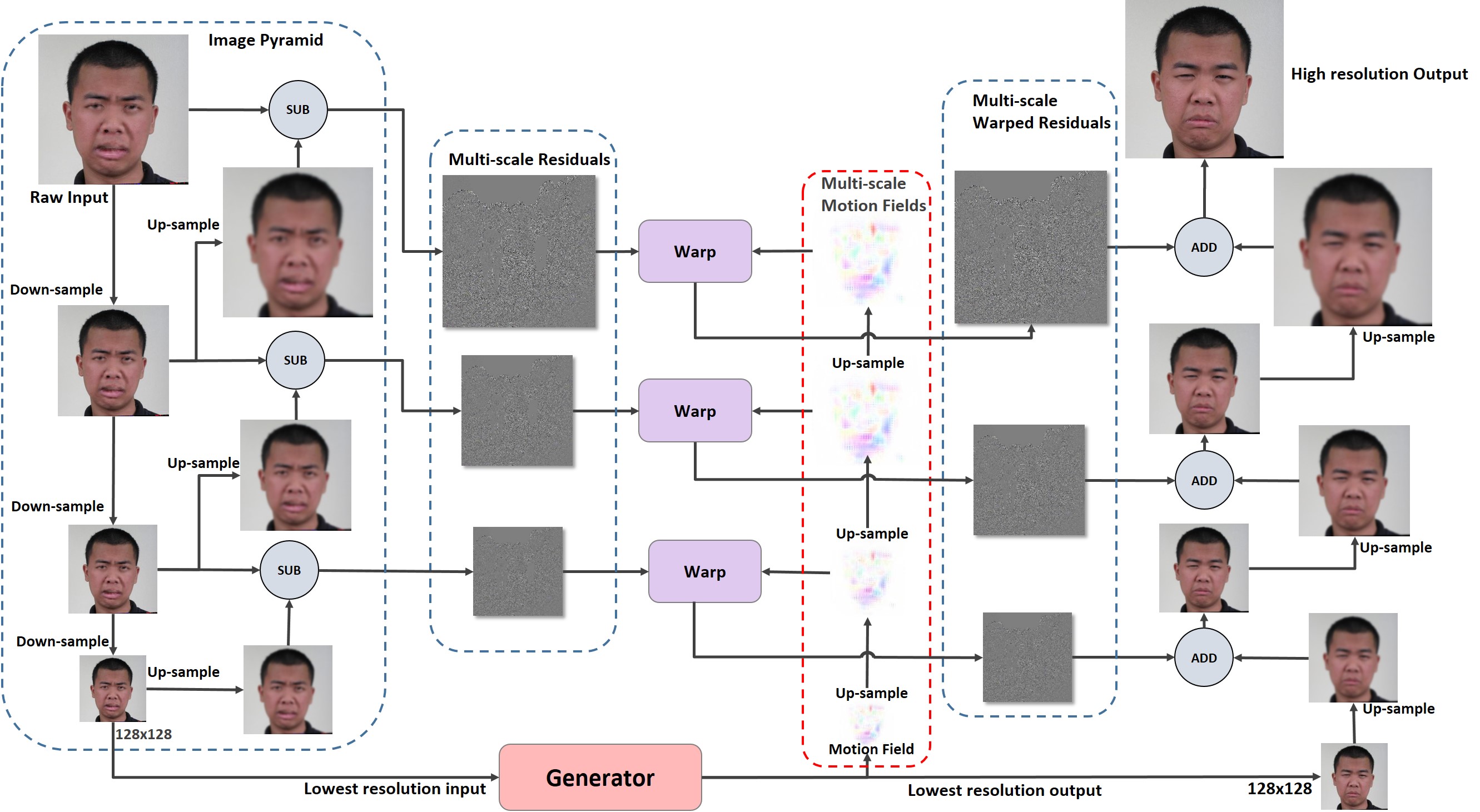}
	\caption{Processing pipeline of multiscale \textbf{ResWarp}.\label{fig:ms-ts}}
\end{figure*}

\subsubsection{Training procedure}
We follow a standard WGAN-GP training procedure \cite{gulrajani2017improved}, in which discriminators and generator are alternatively trained with losses defined in Equation~\ref{eq:dlossw}, \ref{eq:dlossr} and~\ref{eq:gloss}: see Algorithm~\ref{alg:procedure}.

\begin{algorithm}
	\small
	\caption{Training of our Proposed Network}
	\label{alg:procedure}
	\SetAlgoLined
	initialization; \\
	\While{G has not converged}{
		\For{i = 1,...,5}{
			Sample batch images $\mathbf{I}$ from training data;\\
			Randomly sample driving signals $\mathbf{c}_t$;\\
			Get manipulated image $\mathbf{O^r} \leftarrow G(\mathbf{I}, \mathbf{c}_t)$;\\
			Sample a random number $\alpha \in U[0,1]$;\\
			Get interpolation $\mathbf{\hat{I}} \leftarrow (1-\alpha) \mathbf{I} + \alpha \mathbf{O^r}$;\\
			Update the discriminator $\mathbf{D}^r$ with loss $l^{\mathbf{D}^r}$;\\
			Update the discriminator $\mathbf{D}^w$ with loss $l^{\mathbf{D}^w}$;\\
			
		}
		Sample batch images $\mathbf{I}$ from training data;\\
		Randomly sample driving signals $\mathbf{c}_t$;\\
		Get result $\mathbf{O^r} \leftarrow G(\mathbf{I}, \mathbf{c}_t)$;\\
		Update generator G with loss $l_g$;\\
	}
\end{algorithm}

\subsubsection{Data preprocessing/augmentation}

Some prior works (e.g., \cite{pumarola2018ganimation}) only process tightly-cropped faces, which limits the application range. In our experiments, we enlarged facial region and ensure the whole headshot is covered by the image. During training, we also randomly crop/flip the images and adjust color values (e.g., brightness, contrast and hue) to augment the data.

\subsection{Residual warping module}
\label{sect:reswarp}
As demonstrated in Figure~\ref{fig:method} (a), the key idea of \textbf{ResWarp} module is to warp the input residual map $\textbf{R}^i$ to $\textbf{R}^w$ and then add $\textbf{R}^w$ onto $\textbf{O}^u$ to generate a large sharp result. In Figure~\ref{fig:method} (a), the raw input $\textbf{I}^h$ is down-sampled once to $128\times128$ and get only one residual map, which we refer to as vanilla \textbf{ResWarp}. 

An alternative way is to down-sample the raw input multiple times and compute a pyramid of residual maps. As illustrated in Figure \ref{fig:ms-ts}, each time we half the size of the input image by down-sampling and compute a residual map upon images before and after down-sampling. We repeat the process until the image size reaches 128$\times$128. After we obtain the low-resolution animated result from the generator, we up-sampled it by $2\times$ and add a warped residual map to it, and repeat the up-sampling and addition until it reaches the raw size. All residual maps are warped by the up-sampled versions of the same motion field. Note that the up-sampling of the motion field and images are based on bilinear interpolation, and down-sampling is based on neighbor averaging. As the residual maps are warped at multiple scales and the final result is obtained in a coarse-to-fine manner, we call it as multiscale \textbf{ResWarp}. We experimented both variations of multiscale and vanilla \textbf{ResWarp} and report the analysis results in experimental resutls.

\section{Experimental Results}
\subsection{Implementation details}

We evaluate the proposed method on three datasets: CelebA~\cite{liu2015deep}, EmotionNet~\cite{fabian2016emotionet}, and RaFD~\cite{langner2010presentation} \cite{rossler2018faceforensics}. All facial images in these datasets are aligned, cropped and resized to $128\times128$ before used for training. All models are trained on a NVIDIA 1080 Ti GPU with images of resolution 128$\times$128 and batch size set to 10. After training, the models are then tested on unseen images of various resolutions of 512 to 4K on one GPU. The final model has a total of 2.7M$\sim$9.4M parameters, varying by specific datasets. The model is implemented using Pytorch 0.4.1 and TorchVision 0.2.1.

In particular, we trained on CelebA for both attribute-driven and example-based facial animation. For attribute-driven facial animation, we use expression categories as driving signal. Since the CelebA dataset only provides labels of two categories (non-smiling and smiling), we train a model that can drive a given face from non-smile to smile, or vice versa. As expression category is not continuous, we are not able to generate an animation sequence by changing the driving signal. Therefore, we first generate a motion field for the target expression, and then produce the animation sequence by linearly interpolating the motion field \cite{pumarola2018ganimation}. 

To enable animating a face based on exemplars, we use action units (AUs) as driving signal. A third-party AU estimator, OpenFace~\cite{schroff2015facenet}, is employed to generate annotations for training images. Once trained, the model can drive a still face to mimic the target facial action using a continuous sequence of AUs extracted from an exemplar video.

As for EmotionNet and RaFD datasets, they provide abundant categorical expression labels (26 categories for EmotionNet, 8 categories for RaFD). Therefore, we mainly experimented the attribute-driven facial animation on these two datasets. In particular, we trained separate models for each dataset in terms of the application of attribute-driven animating. During inference, one single model for each dataset is able to generate animations of all available expression categories. In other words, one single model can generate facial animations with different expression categories. 

\subsection{Intermediate and final results}

\begin{figure}
  \centering
  \includegraphics[width=.8\linewidth]{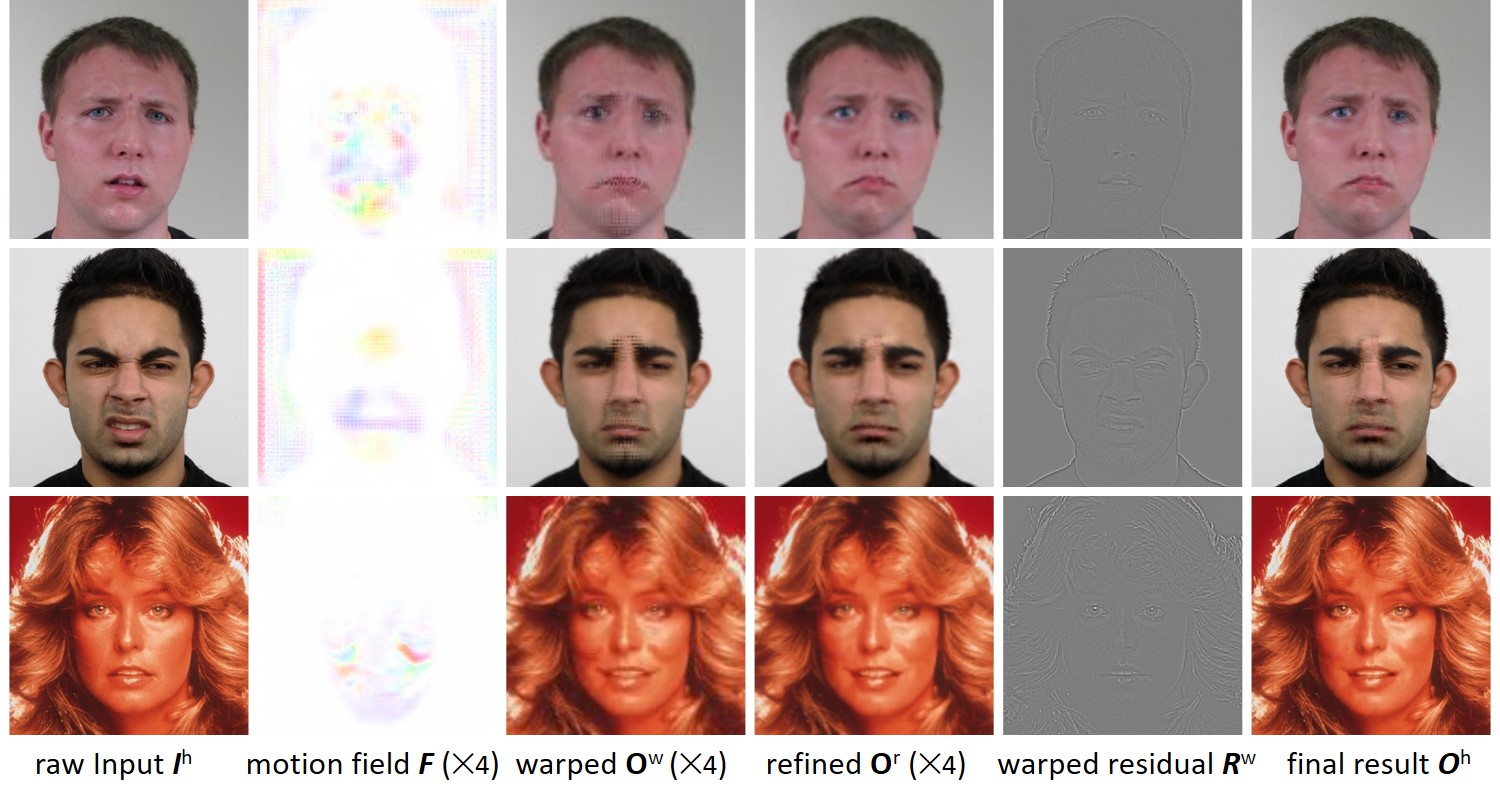}
  \caption{Illustration of intermediate and final results produced with our method. Input size is 512$\times$512. Sizes of $\textbf{F}$, $\textbf{O}^w$ are $\textbf{O}^r$ are originally 128$\times$128, which are up-sampled by 4 times for visualization purpose. Top, middle and bottom row are produced by models trained on EmotionNet, RaFD, CelebA respectively. Driving signal in this setting is expression categories. In particular, source and target expressions from top to bottom are: angrily surprised$\rightarrow$sadly surprised, disgusted$\rightarrow$sad, neutral$\rightarrow$smile. \label{fig:temp}}
\end{figure}

In Figure~\ref{fig:temp}, we visualize learned motion fields $\textbf{F}$, coarse warping results $\textbf{O}^w$ and refined results $\textbf{O}^r$,  warped residual maps $\textbf{R}^w$ and final outputs $\textbf{O}^h$. Figure~\ref{fig:temp} demonstrates that our Motion Network produces reasonable motion field and warping results. However, without refinement, the warping results may have distortions and artifacts, and suffer lack of details in certain regions. The refined results given by the Refinement Network look more realistic and have less artifacts. In summary, the Motion Network aims to generate reasonable geometric deformation, e.g. lips stretched for smiling faces, and is able to tell different parts of faces required to be deformed to achieve target facial expression. The Refinement Network further suppresses noise and artifacts, and adds more textures on warping results to make images look more realistic. The ResWarp module transfers the motion field to residuals. The final result is obtained by adding the warped residual map $\textbf{R}^w$ to the up-sampled refined result $\textbf{O}^u$. Note that all results shown in Figure \ref{fig:temp} are produced with input size of 512$\times$512.

\begin{figure}
  \centering
  \includegraphics[width=.8\linewidth]{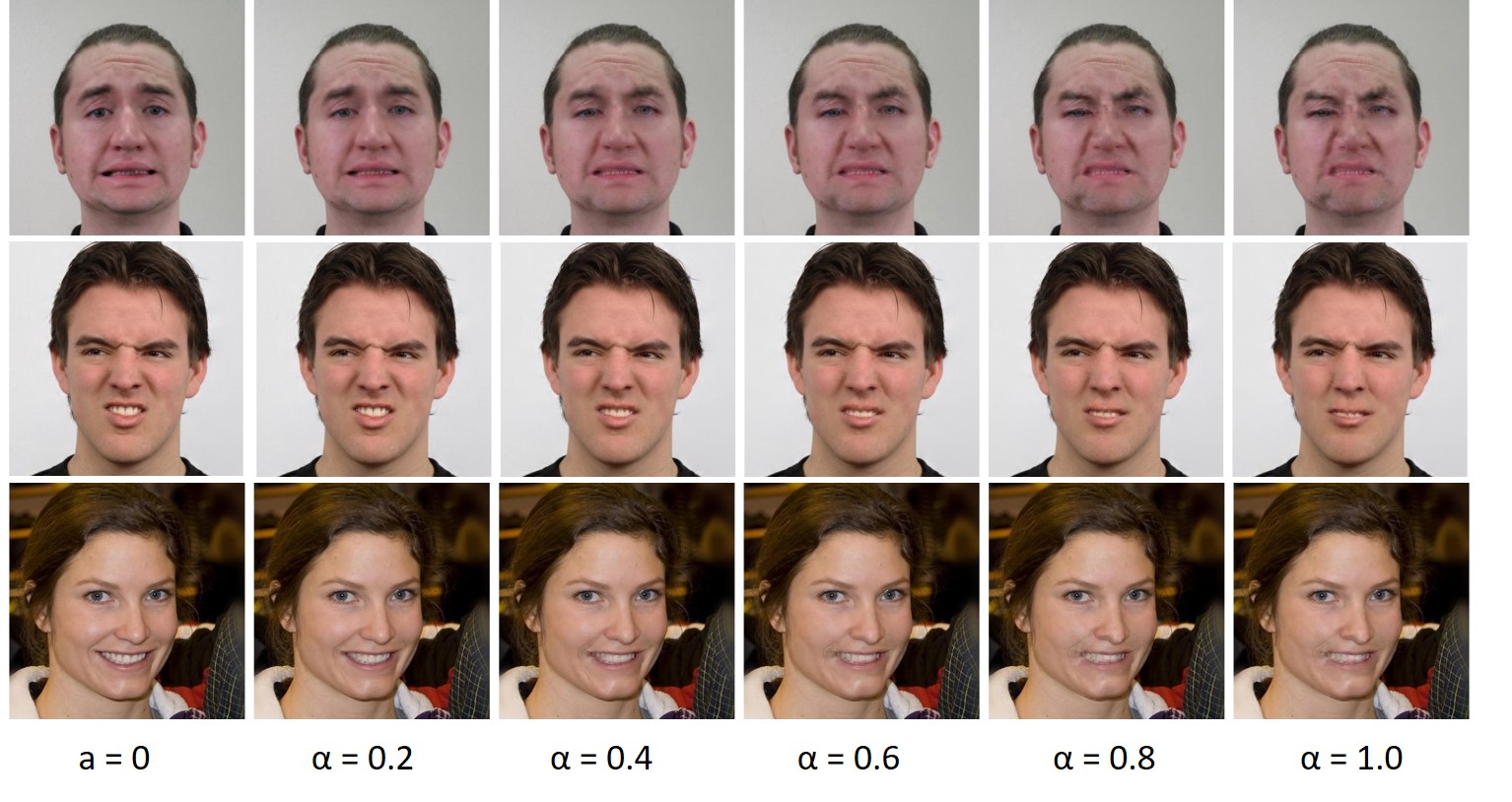}
  \caption{Attribute-driven animation results, produced by linearly interpolating the motion fields using factor $\alpha$ = $0.0$, $0.2$, $0.4$, $0.6$, $0.8$, $1.0$. Source and target expressions from top to bottom are: sadly fearful$\rightarrow$appalled, disgusted$\rightarrow$angry, smile$\rightarrow$neutral. \label{fig:seq}}
\end{figure}

\begin{figure}
  \centering
  \includegraphics[width=.8\linewidth]{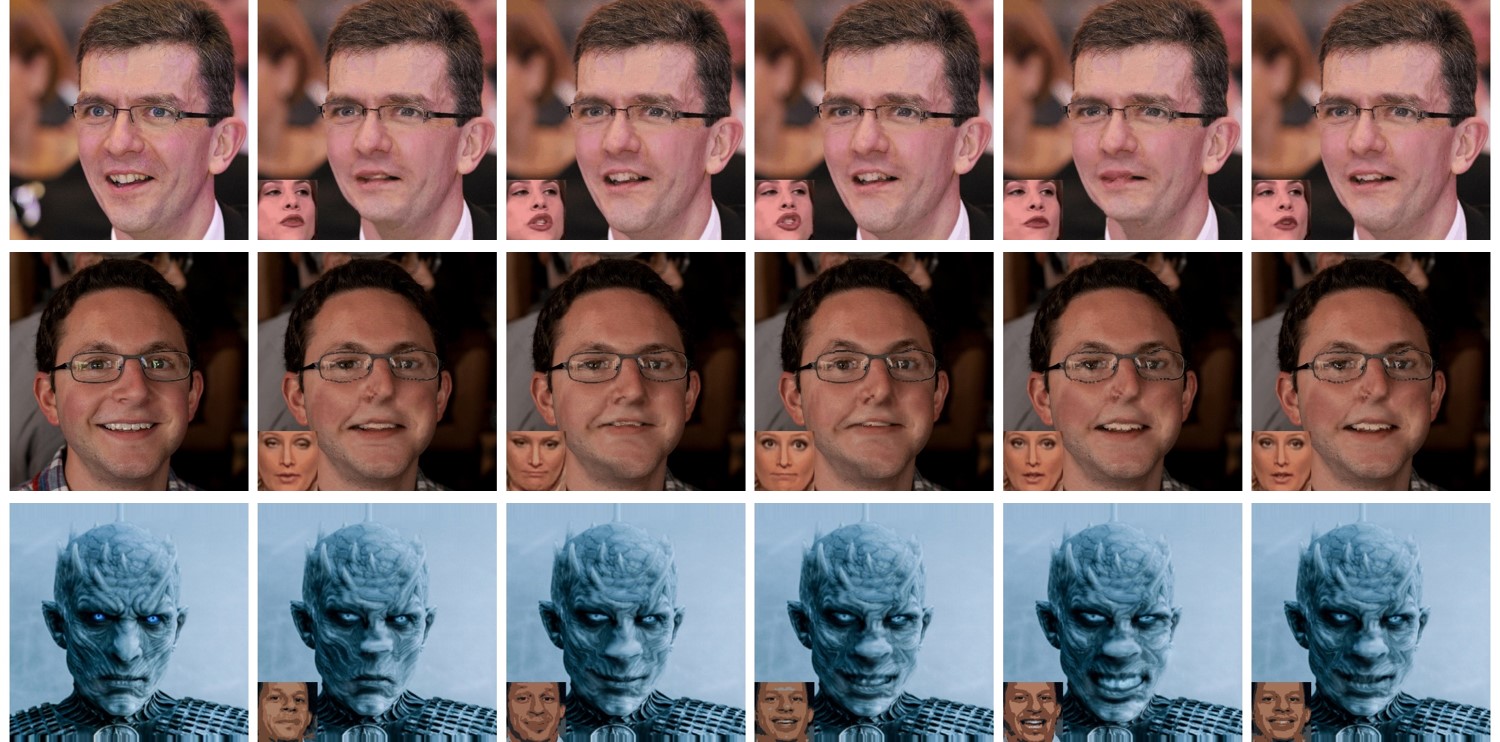}
  \caption{Example-based animation results, produced by different driving videos as examples.\label{fig:seq_au}}
\end{figure}

In addition, Figure \ref{fig:seq} illustrates results of attribute-driven facial animation, in which the animation sequence is generated by linearly interpolating the motion field. Figure \ref{fig:seq_au} illustrates the example-based animation results by using the exemplar AUs as driving signals. Figure~\ref{fig:teaser} demonstrates high-resolution animating results of sizes ranging from 512 to 4K. More animation results are provided in the supplementary materials.

\subsection{Comparisons with facial reenactment methods}

We compare our method with state-of-the-art methods including Ganimation~\cite{pumarola2018ganimation}, x2face \cite{wiles2018x2face} and few-shot vid2vid\cite{zakharov2019few} for the application of example-based facial animation, where an input image is animated based on a driving video. To guarantee fairness, we use the same sets of 100 image-video pairs, in which the 100 images are randomly selected from FFHQ datasets \cite{karras2019style} and driving videos are randomly selected from the original split of FaceForensics dataset \cite{rossler2018faceforensics}. Note that the input images have size of 1024$\times$1024 and are unseen for all contesting models.

As the pretrained Ganimation model officially provided was trained on tightly-cropped faces of size 128$\times$128, we crop the face regions from the input images, resizes them to 128$\times$128 before testing. Once processed, we resize the animated faces to original sizes and paste them back to the original image. The official pretrained model of x2face was trained on VoxCeleb video dataset with image size of $256\times256$. Few-shot vid2vid is trained on FaceForensics \cite{rossler2018faceforensics} with image size of 128$\times$128. As the official pretrained models of vid2vid \cite{zakharov2019few} are not provided, we use their official sources codes to train a new model. Limited to GPU memory, we can only train the model on 128$\times$128 images with a single GTX TITAN 1080 ti GPU which has total memory of 11 GB. Our \textbf{ATW} model is trained on CelebA dataset with image size of 128$\times$128 using AU as driving signal, and it can do inference on the 1K test images. Note that few-shot vid2vid and x2face can do either one-shot or many-shot facial animation, but here we only evaluate the task of one-shot animation.

The driving signals of Ganimation, x2face and few-shot vid2vid are different, i.e., action units, driving vectors extracted from RGB images, and edge maps that connect facial landmarks, respectively. We extract those information from the driving videos using either external tools (OpenFace \cite{schroff2015facenet} for AU, Face-Alignment \cite{bulat2017far} for facial landmarks) or utilities provided by itself (the driving network provided in \cite{wiles2018x2face} for driving vectors). 

\subsubsection{Qualitative comparison}
\begin{figure}
	\centering
	\includegraphics[width=.8\linewidth]{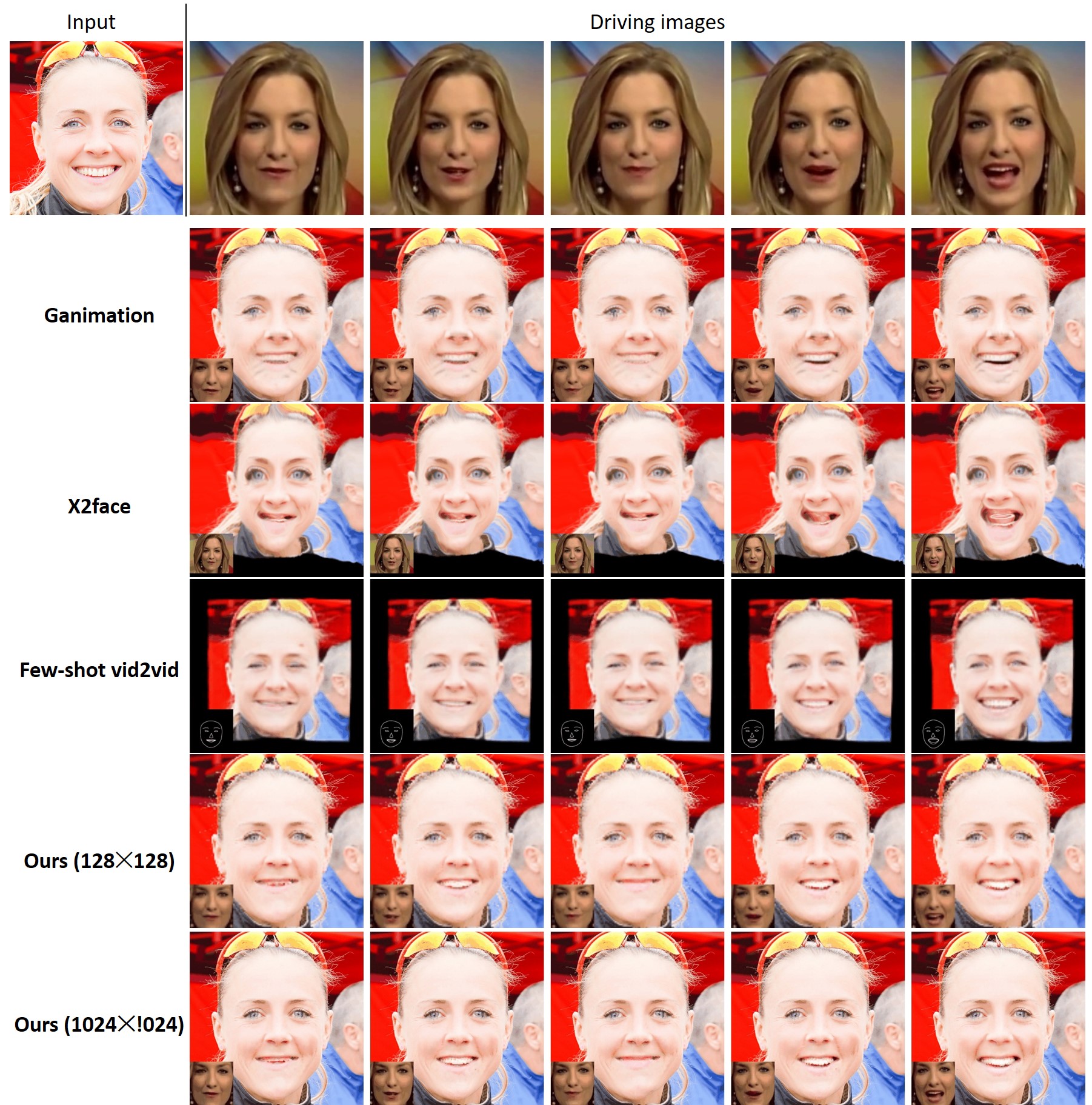}
	\caption{Comparison against the state-of-the-arts for example-based facial animation. All results are generated with the same input image-video pair and resized to 1024$\times$1024 for fair comparison. Driving signal for each resulting image is visualized at left bottom corner. Please zoom in to see more details. \label{fig:compare}}
\end{figure}

The visual comparisons between all contesting methods are shown in Figure \ref{fig:compare}. We find that our method produces the sharpest and the most realistic results. At the same time, the identity of generated faces are well preserved. X2face generates unrealistic expressions, while Ganimation produces visible artifacts at the mouth regions. Few-shot vid2vid produces realistic expressions but fails to preserve accurate input identity.

\begin{figure}
	\centering
	\includegraphics[width=.8\linewidth]{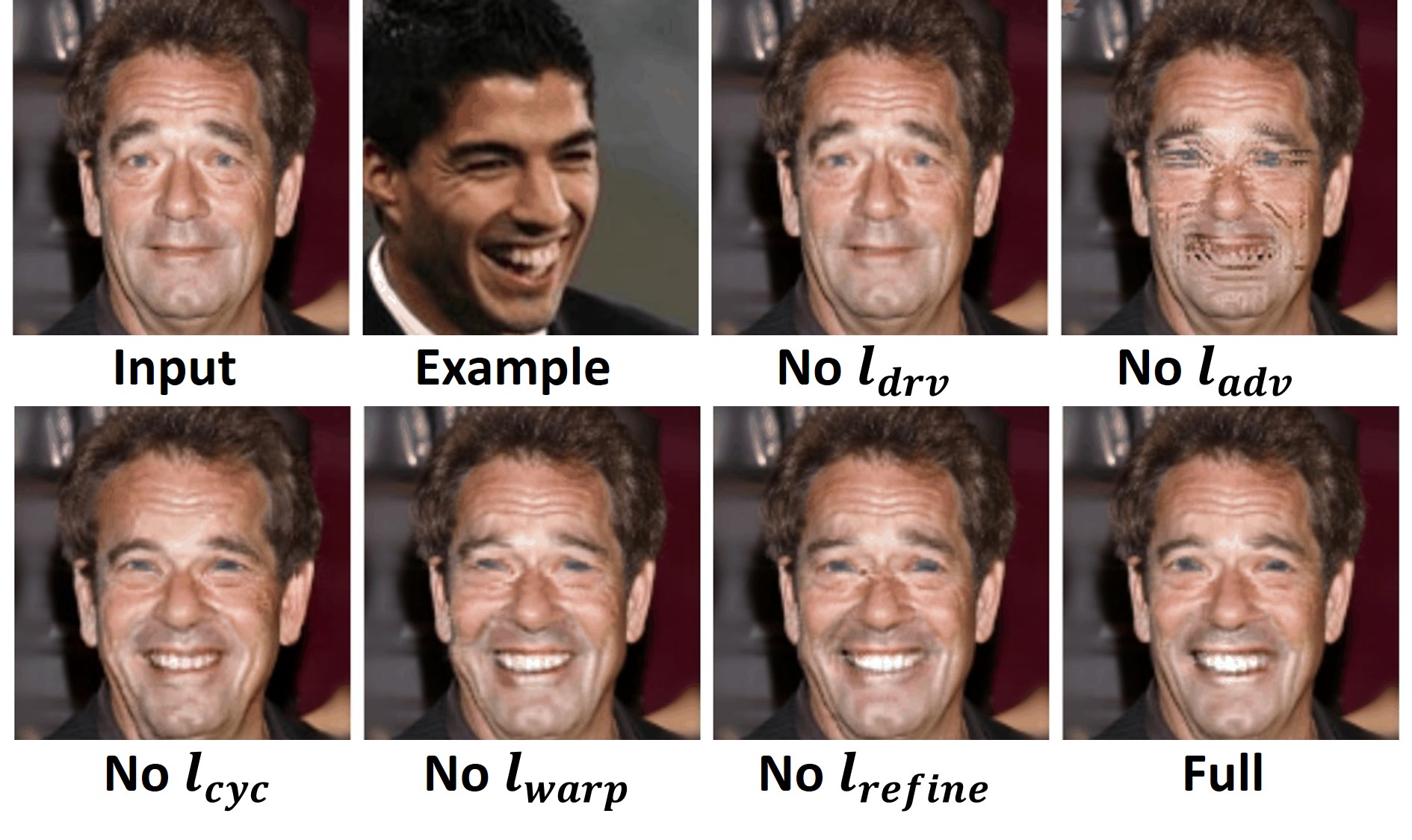}
	\caption{Ablation studies on training losses in the task of example-based facial animation. E.g., ``No $l_{cyc}$'' indicates that $l_{cyc}$ term was removed, and ``Full'' indicates all loss terms were added.\label{fig:loss}}
\end{figure}

\begin{table*}
	\centering
	\caption{Evaluation results on 100 image-video pairs. Note that memory consumption and time cost are tested on a framewise basis. \label{table:compare}}
	\tiny\addtolength{\tabcolsep}{-1pt}
	\begin{tabular}{c||cccc|cccc|c|cc}
		& \multicolumn{4}{|c}{Model Details} &  \multicolumn{5}{|c}{Quality}  & \multicolumn{2}{|c}{Efficiency}  \\
		Method & train dataset & driving signal &  train size & test size& FID & ID CSS & FWE  &  AU Err.& Human &  Mem & time\\
		\hline
		\hline
		Ganimation\cite{pumarola2018ganimation} &  CelebA  &  AU & 128$\times$128    &  128$\times$128  & 114.6 & \textbf{0.798} & 12.5    &  8.1   & 19.0\%   & 653MB & 16.8ms \\
		Few-shot vid2vid\cite{zakharov2019few}   & FaceForensics & edge map & 128$\times$128   & 128$\times$128 & 146.7 & 0.675 & 9.4  & 6.4    & 26.6\%  & 1559MB& 29.3ms\\
		x2face \cite{wiles2018x2face} & VoxCeleb &   embedding & 256$\times$256   &   256$\times$256  &126.8   & 0.702 & 12.0 &    10.4 &   19.6\%     &   1079MB&  4.1ms  \\
		\textbf{Ours}  &  CelebA  &  AU & 128$\times$128   & \textbf{1024$\times$1024}  &   \textbf{107.0}&  0.776 &  \textbf{6.2}  &   \textbf{4.5}   &  \textbf{34.8\%} & 669MB & 21.3ms
	\end{tabular}
\end{table*}

\subsubsection{Quantitative comparison}
\label{sect:quat}
We use four objective metrics to measure the quality of generated animations. Firstly, FID \cite{heusel2017gans} is used to measure feature distribution discrepancy between real and generated faces. Secondly, the correctness of generated expressions is evaluated by comparing the AU features of driving faces with such attributes in generated faces. Specifically, the mean value of the sum of squared errors between the AU vector of a generated expression and that of the corresponding driving expression is used to indicate correctness of expression imitation. Thirdly, we use Flow Warping Error (FWE) as defined in \cite{lai2018learning} to measure the temporal coherency of generated animations, in which FlowNet2 \cite{ilg2017flownet} is used as the optic flow estimator. At last, we employed a pretrained face recognizer VGGFace \cite{parkhi2015deep} to assess the preservation of facial identity in animated sequences. Specifically, the activations of the second last layer (bottleneck layer) of VGGFace model is used as the embedding of face identities. The embedding vector of the input face is compared with that of each animated face and a cosine similarity score (CSS) between them is computed. Mean of CSS of all animated faces is used as the indication of identity preservation (the higher the better).

Results in Table \ref{table:compare} demonstrate that our framework achieves the lowest FID score among all methods, implying that the distribution of our generated images is closest to that of real images. In addition, the lowest Flow Warping Error and AU Error achieved by our method indicates the effectiveness of our method in preserving the temporal coherency of generated animation sequences and imitating the target expression. Both Ganimation and our model achieve high identity similarity and do well in preserving face identities. In terms of test efficiency, our method costs comparable time and memory as competitors while generating significantly higher-resolution results.

\begin{figure}
	\centering
	\includegraphics[width=.8\linewidth]{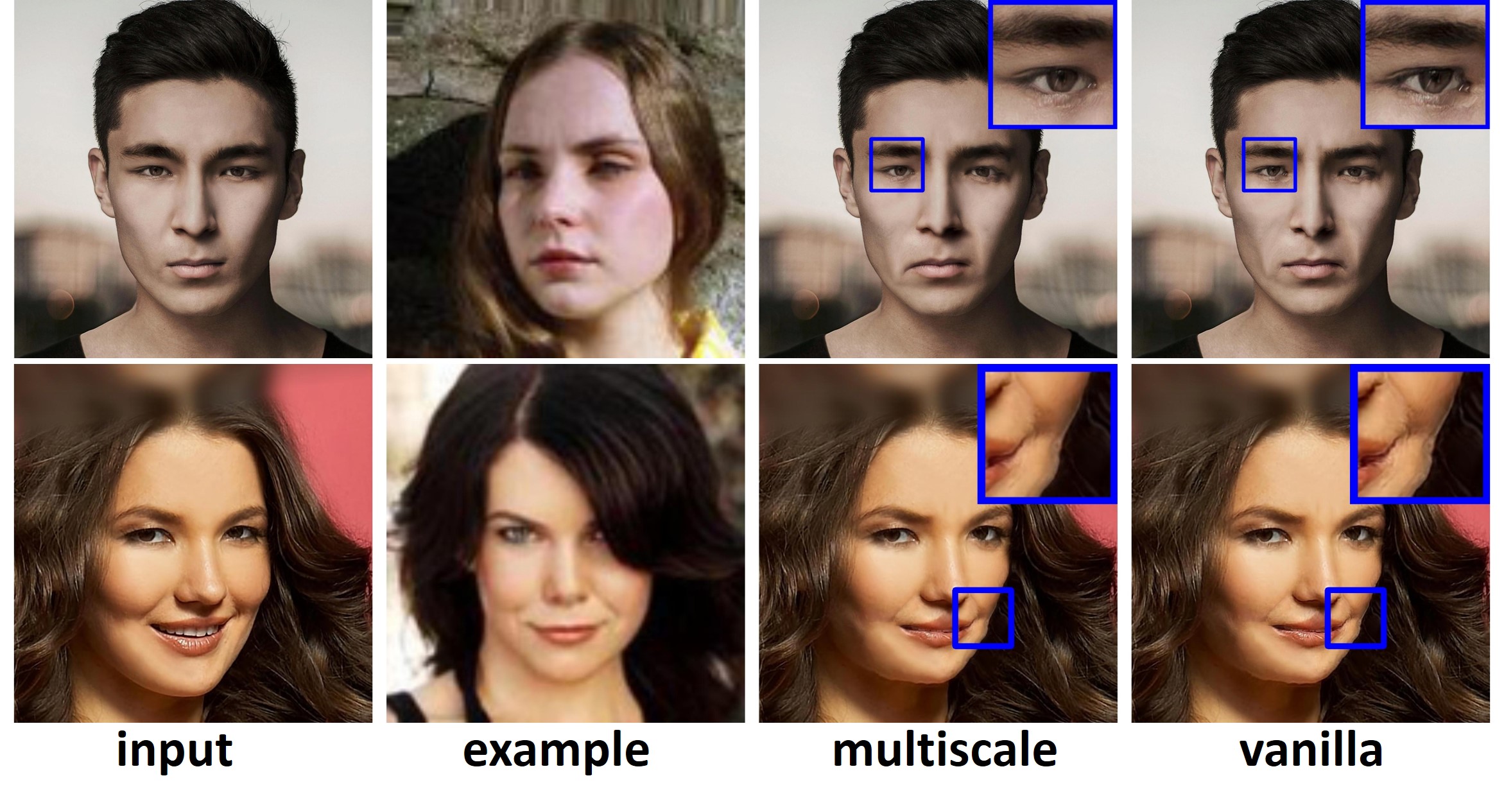}
	\caption{Comparisons of multiscale \textbf{ResWarp} against vanilla \textbf{ResWarp} for example-based facial expression animation. The multiscale \textbf{ResWarp} appears to produce more realistic textures.\label{fig:multiscale}}
\end{figure}

We also conduct user study for method comparison among 100 randomly selected image-AU pairs. Given an input image, 20 Amazon turks are instructed to choose the best item based on quality of generated animation, perceptual realism, and preservation of identity. Then the percentage of ratings that favor each method is computed and compared: see Table \ref{table:compare}. The human assessment results demonstrate that our method obtains the highest average rating among different facial animation methods.

\subsection{Ablation Study}
\subsubsection{Analysis of training losses}
We carried out ablation studies on the training losses for the generator. We trained the generator with a specific loss term removed from Eq.~\ref{eq:gloss}, and see how the results are altered, as shown in Figure \ref{fig:loss}. We observe that without driving loss, the generated image is almost identical to the input and fails to achieve what is desired by the driving signal. Without adversarial loss, the result suffers severe artifacts. Without cycle loss, the identity is less preserved. Without refine loss and warp loss, some minor artifacts become visible: refer to the eyes in Figure \ref{fig:loss}. We also conducted quantitative evaluations as shown in Table \ref{table:compare_loss}, which further verifies the property of each loss term. The proposed refine loss contributes to the improvements of the temporal coherency and image quality. In practice, we recommend careful tuning of hyper-parameters based on specific requirements.

\begin{table}
	\centering
	\caption{Quantitative evaluation results on 100 image-video pairs when using different configurations of training losses. Drawbacks of each configuration are highlighted in red color.\label{table:compare_loss}}
\begin{tabular}{c||cccc}
Method & FID & ID CSS & FWE  &  AU Err. \\
	\hline
	\hline
No $l_{drv}$  & \textbf{0.2}&  \textbf{0.99} &  \textbf{0.6}  &   {\color{red} 9.8}  \\
No $l_{adv}$  & {\color{red}146.7}&  {\color{red}0.512} &  {\color{red}7.3}  &   {\color{red}5.7}  \\
No $l_{cyc}$   &   106.2&  {\color{red}0.714} &  6.2  &   \textbf{4.1}  \\
No $l_{ref}$   &   {\color{red}112.4}&  0.773 &  {\color{red}6.9}  &   4.4  \\
No $l_{warp}$   &   {\color{red}109.1}&  0.782 &  6.0  &   {\color{red}4.8}  \\
\textbf{Full}  &   107.0&  0.776 &  6.2  &   4.5  \\
\end{tabular}
\end{table}

\subsubsection{Analysis of multiscale and vanilla \textbf{ResWarp}}
As mentioned in Sect. \ref{sect:reswarp}, we experimented two variations of \textbf{ResWarp}: multiscale \textbf{ResWarp} and vanilla \textbf{ResWarp}. Multiscale \textbf{ResWarp} requires computation of a Laplacian Pyramid and warps the residuals at multiple scales, thus being a bit less efficient than vanilla \textbf{ResWarp}. In terms of quality, we observe not much difference between them for images that less than 1024x1024. However, when the image resolution gets larger, multiscale \textbf{ResWarp} generates visually better results: see Figure~\ref{fig:multiscale}, especially the left eye of the young man in the first row. This matches our intuition that the vanilla \textbf{ResWarp} may fail to capture enough high frequency details to improve the final target image, especially for high resolution images. In terms of efficiency, the multiscale \textbf{ResWarp} is about 2ms slower than vanilla \textbf{ResWarp} when doing inference on 1K images with our machine (Intel Core i7-8700K CPU@3.70GHz). In practice, one needs to choose the appropriate balance between the final quality and computational efficiency. 

\subsection{More experimental results}
More experimental results including comparisons of \textbf{ResWarp} and super-resolution methods, comparison of our two-stage generator with other warping-based generators, ablation studies on the design of the \textbf{ResWarp} module and failure examples of our method are provided in the supplementary materials.

\section{Conclusion}

We presented a novel \textbf{A}nimating-\textbf{T}hrough-\textbf{W}arping (\textbf{ATW}) framework, which enables animating HD facial images with limited resources. We experimented our method on three datasets and two applications (attribute-driven and example-based facial expression animation), verifying its effectiveness and efficiency. So far, our method is the only neural model that can successfully animate an ultra-high-resolution image. Qualitative and Quantitative comparisons with existing methods shows our method generally improves the temporal coherency and image quality in generated animation sequences without compromising efficiency, in the application of one-shot facial expression animation.

However, our method also suffers some failure examples, especially when generated faces involve synthesis of novel contents (e.g., teeth). Some typically failures are provided in the supplementary material. So far, we only experimented our method for animation of facial expressions. In the future, we intend to extend our method for reenactment of head poses and even arbitrary genres of objects or scenes.

\section*{Acknowledgment}
We want to acknowledge Rui Ma, Daesik Jang, James Gregson, Shao Hua Chen, Kevin Cannons and Peng Deng in Huawei Technologies for their help and support in the project.

\bibliographystyle{splncs}
\bibliography{references}

\appendix

\clearpage
\section{Appendix}
\subsection{Comparisons against super-resolution methods}

\begin{figure}
	\centering
	\includegraphics[width=.8\linewidth]{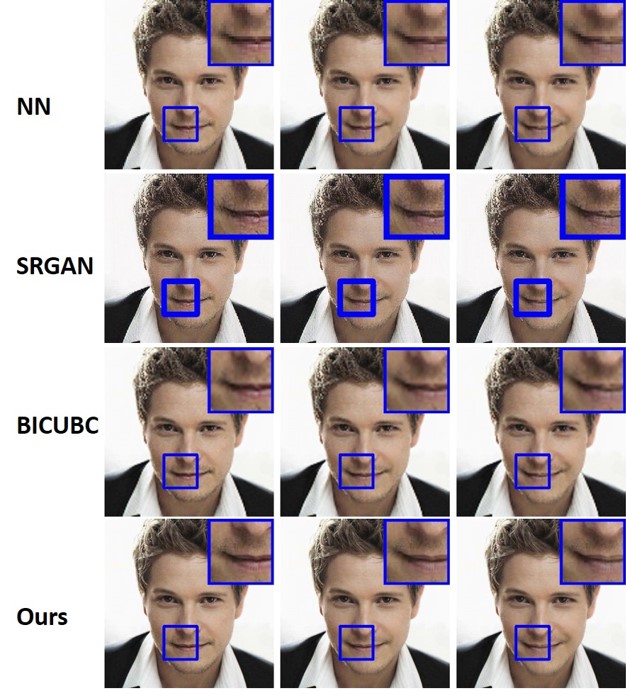}
	\caption{Comparisons of \textbf{ResWarp} against super-resolution methods. NN refers to Nearest Neighbor interpolation. We observe that results by NN contain tiling artifacts, and those produced by Bicubic interpolation suffer lack of sharpness. Results by SRGAN are sharp but look unnatural. Please zoom in to see more details. \label{fig:sr}}
\end{figure}

The \textbf{ResWarp} module in our framework  is responsible for constructing a high-resolution result from a low-resolution result of $128\times128$, which can also be achieved with super-resolution methods. We compared our final results $\textbf{O}^h$ against those produced by super-resolution methods from the same low-resolution results $\textbf{O}^r$, as shown in Figure \ref{fig:sr}. Note that SRGAN~\cite{ledig2017photo} is a learning-based method that can perform 4$\times$ super-resolution and the official pretrained model was trained on DIV2K. we observe that the textures and details recovered with our \textbf{ResWarp} module look more realistic than super-resolution methods such as SRGAN \cite{ledig2017photo}, or naive up-sampling with Nearest Neighbor and Bicubic interpolation. We also conducted quantitative evaluations of these methods on the same 100 image-video pairs as used in Section \ref{sect:quat}: see Table \ref{table:compare_sr}. Since these methods share the same low-frequency components, we do not need to compare the AU Error and identity preservation. Quantitative results show \textbf{ResWarp} achieves the lowest FID and best temporal coherency.

\begin{table}
	\centering
	\caption{Quantitative evaluation results on 100 image-video pairs when using different super-resolution methods. \label{table:compare_sr}}
	\begin{tabular}{c||cc}
		Method &  FID & FWE \\
		\hline
		\hline
		NN  &  118.2  & 7.9 \\
		BICUBIC & 115.3 & 6.4 \\
		SRGAN  &   110.8 & 7.2   \\
		\textbf{ResWarp}  &  \textbf{107.0} &  \textbf{6.2} \\
	\end{tabular}
\end{table}

\subsection{More Ablation Study}

\subsubsection{Analysis of up-sampling techniques in \textbf{ResWarp}}
\begin{figure}
	\centering
	\includegraphics[width=.8\linewidth]{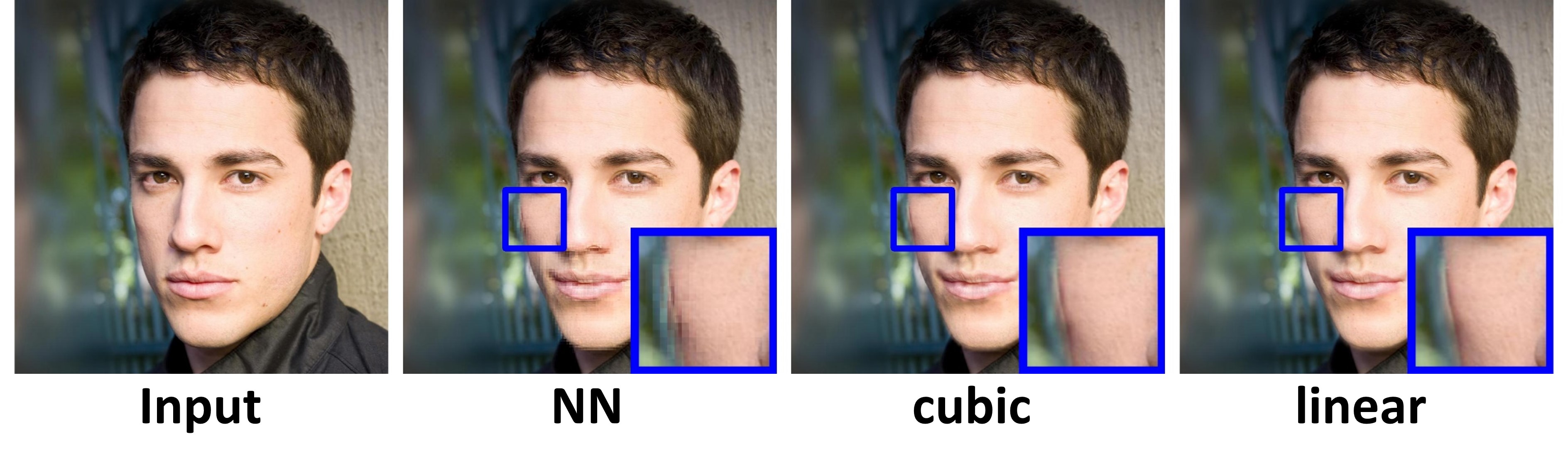}
	\caption{Comparisons of various up-sampling techniques used in \textbf{ResWarp} module in the task of attribute-driven facial animation (only last frame is shown). Note that NN refers to Nearest Neighbor interpolation. Differences of results are highlighted with blue boxes.\label{fig:resizing}}
\end{figure}

As in the \textbf{ResWarp} module, one down-sampling and two up-sampling operations are performed: see Figure \ref{fig:method} (a). For down-sampling, we evenly split the raw input image into 128$\times$128 patches and average each patch to form a new pixel. For up-sampling, we experimented three techniques: Nearest Neighbor (NN), Bilinear and Bicubic. Figure \ref{fig:resizing} shows how the choose of up-sampling technique affects the final results. Given the same input image and driving signal, we highlight differences between results generated using these up-sampling techniques. We observed that NN tends to generate tiling artifacts, and Bicubic generates some minor artifacts at boundaries. Bilinear produces the most pleasing results. In our final implementation, we choose bilinear up-sampling technique in the \textbf{ResWarp} module.

\subsection{Failure examples}
\begin{figure}
	\centering
	\includegraphics[width=.8\linewidth]{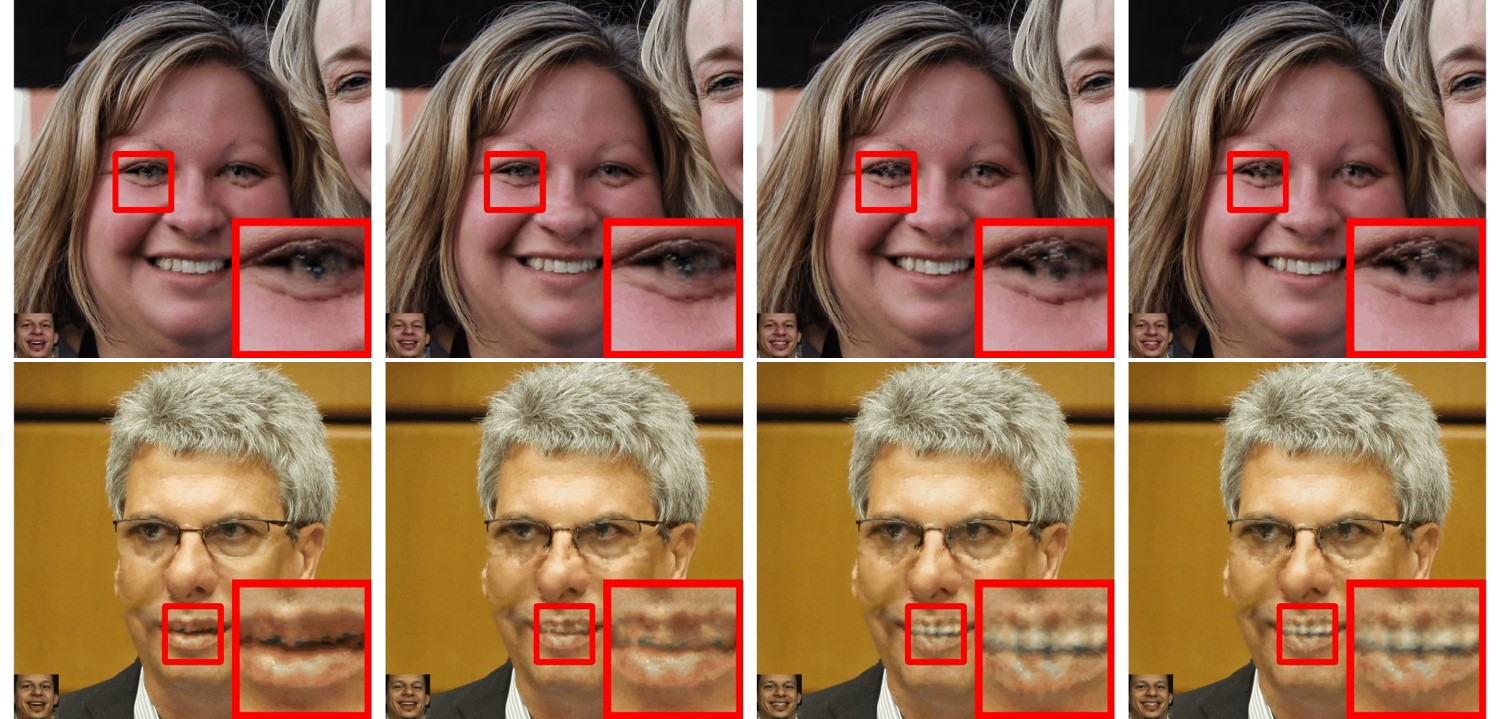}
	\caption{Failure examples of our method. Sizes of raw input images are 1024$\times$1024. \label{fig:failure}}
\end{figure}

Figure \ref{fig:failure} shows some typical failure examples of our method. When the results involve synthesis of novel contents (e.g., teeth when opening mouth), our method is prone to suffer blurriness as the corresponding residuals cannot be warped from the source image: see the bottom row of Figure \ref{fig:failure}. In addition, our method occasionally predicts invalid motion field and causes motion artifacts: see the top row of Figure \ref{fig:failure}.

\subsection{Combination of prior warping-based model and \textbf{ResWarp}}
\begin{figure}
	\centering
	\includegraphics[width=.8\linewidth]{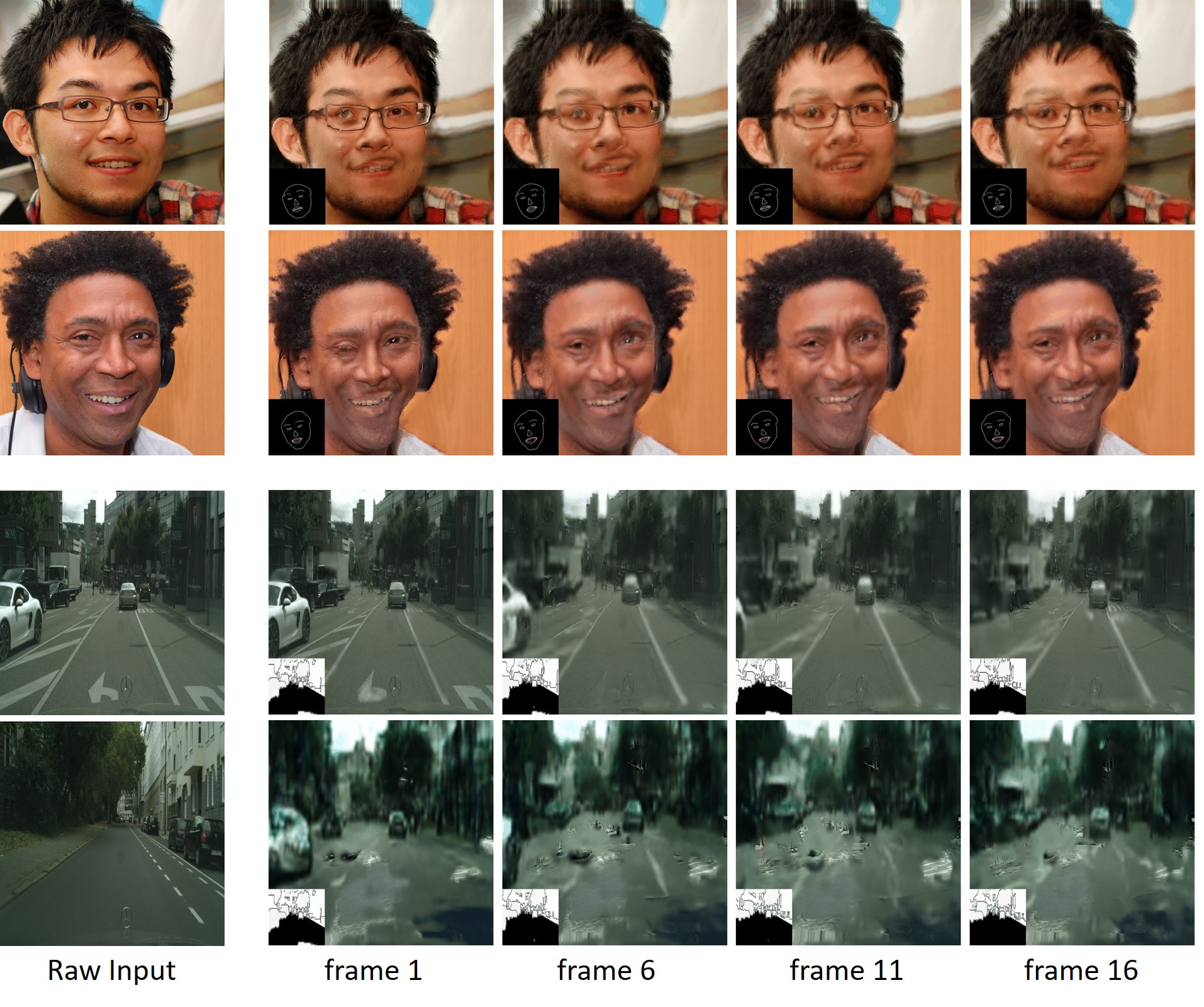}
	\caption{Results of combining few-shot vid2vid and \textbf{ResWarp}. Top two rows are generated by combining the few-shot vid2vid model trained on FaceForensics~\cite{rossler2018faceforensics} and \textbf{ResWarp} module, and bottom two are produced by combining the model trained on CityScape~\cite{cordts2016cityscapes} and \textbf{ResWarp}. Sizes of all input images are 1024$\times$1024. \label{fig:combine}}
\end{figure}

Since existing warping-based models such as \cite{siarohin2019animating,endo2019animating,zakharov2019few} also predict a motion field that warps the input to its output, combining them with our proposed \textbf{ResWarp} module could enable their methods applicable on high resolution images as well. That is to say, we can use their models to generate a low-resolution result and the motion field, and then employ our \textbf{ResWarp} module to warp the raw input residuals and produce high-resolution results. We experimented the combination of few-shot vid2vid and \textbf{ResWarp} module on tasks of animating faces and street views: see Figure \ref{fig:combine}. As shown in Figure \ref{fig:combine}, some results suffer the misalignment of the residual map and low-frequency component (see the lip and glasses in 1st row of Figure \ref{fig:combine}), and some other results suffer severe regional or global blurriness (2nd-4th rows of Figure \ref{fig:combine}). The reason is that the few-shot vid2vid model is not well aligned with the \textbf{ATW} framework, and the motion field and the warped result in their model do not play a dominant role in the generation of the low-frequency component. 

\end{document}